\documentclass[sigconf]{acmart}

\renewcommand\footnotetextcopyrightpermission[1]{} % removes footnote with conference information in first column
\makeatletter
\renewcommand\@formatdoi[1]{\ignorespaces}
\makeatother

\usepackage{booktabs} % For formal tables
\usepackage{soulutf8}
\setcopyright{rightsretained}
\usepackage{xspace}
\newcommand*{\eg}{\emph{e.g}.\@\xspace}
\newcommand*{\ie}{\emph{i.e}.\@\xspace}
\newcommand*{\etc}{\emph{etc}.\@\xspace}
\newcommand*{\etal}{\emph{et al}.\@\xspace}

\begin{document}
\title{Temporal Multimodal Fusion \\for Video Emotion Classification in the Wild}

\author{Valentin Vielzeuf}
\affiliation{%
    \institution{Orange Labs}
  \city{Cesson-S\'evign\'e} 
  \state{France} 
  \postcode{35512}
}
\additionalaffiliation{%
    \institution{Normandie Univ., UNICAEN, ENSICAEN, CNRS}
  \city{Caen} 
  \state{France} 
}
\email{valentin.vielzeuf@orange.com}

\author{St\'ephane Pateux}
\affiliation{%
    \institution{Orange Labs}
  \city{Cesson-S\'evign\'e} 
  \state{France} 
  \postcode{35512}
}
\email{stephane.pateux@orange.com}

\author{Fr\'ed\'eric Jurie}
\affiliation{%
    \institution{Normandie Univ., UNICAEN, ENSICAEN, CNRS}
  \city{Caen} 
  \country{France}}
\email{frederic.jurie@unicaen.fr}

\begin{abstract}
This paper addresses the question of emotion classification. The task consists in predicting emotion labels (taken among a set of possible labels) best describing the emotions contained in short video clips. Building on a standard framework  -- lying in describing videos by audio and visual features used by a supervised classifier to infer the labels --  this paper investigates several novel directions.  
First of all, improved face descriptors based on 2D and 3D Convolutional Neural Networks are proposed. Second, the paper explores several fusion methods, temporal and multimodal, including a novel hierarchical method combining features and scores. In addition, we  carefully reviewed the different stages of the pipeline and designed a CNN architecture adapted to the task; this is important as the size of the training set is small compared to the difficulty of the problem, making generalization difficult. The so-obtained model ranked 4th at the 2017 Emotion in the Wild challenge with the accuracy of 58.8~\%.
\end{abstract}
%  \begin{CCSXML}
% <concept_id>10010147.10010178.10010224.10010225</concept_id>
% <concept_desc>Computing methodologies~Computer vision tasks</concept_desc>
% <concept_id>10010147.10010257.10010293.10010294</concept_id>
% <concept_desc>Computing methodologies~Neural networks</concept_desc>
% <concept_id>10010147.10010178.10010224.10010240</concept_id>
% <concept_desc>Computing methodologies~Computer vision representations</concept_desc>
% <concept_significance>300</concept_significance>

% \ccsdesc[500]{Computing methodologies~Computer vision tasks}
% \ccsdesc[500]{Computing methodologies~Neural networks}
% \ccsdesc[300]{Computing methodologies~Computer vision representations}


\maketitle

\begin{figure}
\includegraphics[width=\linewidth]{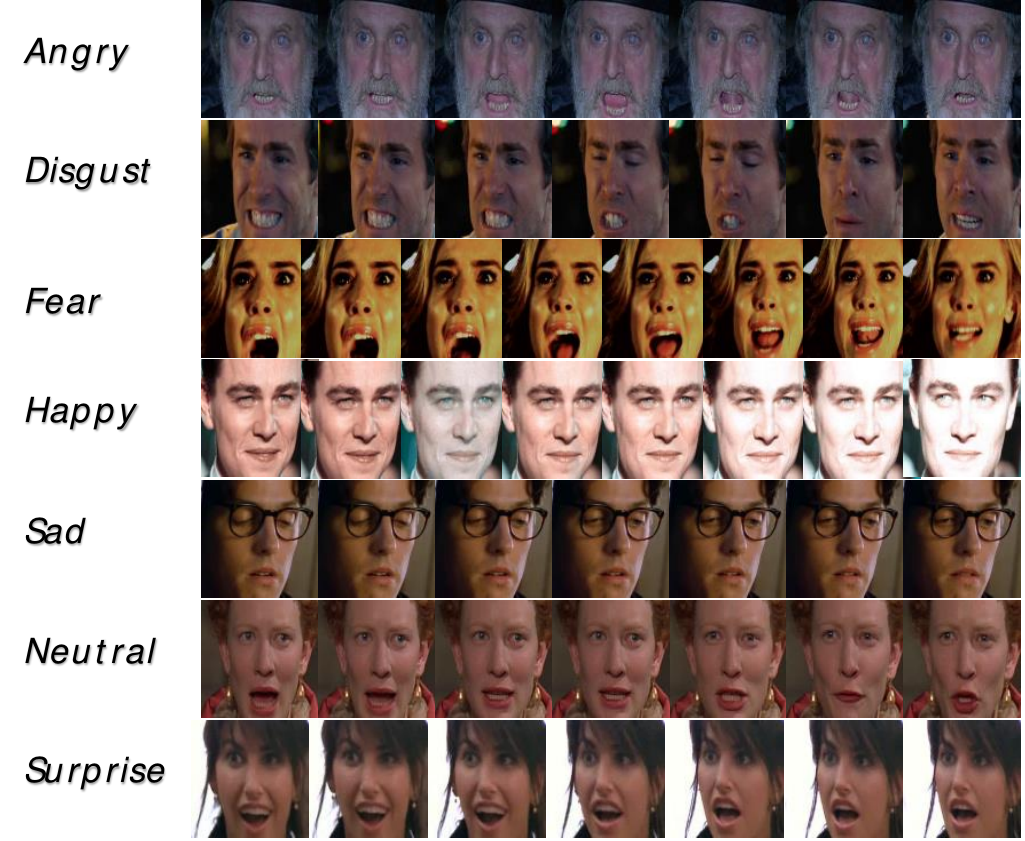}
\caption{The 7 emotions of the AFEW dataset: each row represents an emotion with a set of faces sampled across a representative video clip. Please note that if most of the video clips do contain faces, some of them don't.}
\end{figure}

\section{Introduction and Related Work \label{sec:intro}}

Emotion recognition is a topic of broad and current interest, useful for many applications such as advertising~\cite{kolakowska2013emotion} or psychological disorders understanding~\cite{washington2016wearable}. It is also a topic of importance for research in other areas \eg,  video summarization~\cite{xu2016heterogeneous} or face normalization (expression removal). Even if emotion recognition could appear to be  an almost solved problem in laboratory controlled conditions, there are still many challenging issues in the case of videos recorded in the wild. 

This paper focuses on the task of emotion classification in which each video clip has to be assigned to one and only one emotion, based on its audio/video content. The classes are usually the six basic emotions \ie, anger, disgust, fear, happiness, sadness and surprise, in addition to the  neutral class, as in the {\em Audio-video Emotion Recognition} sub-task of the Emotion Recognition in the Wild Challenge \cite{dhall2012collecting}. More precisely, this paper presents our methodology, experiments as well as the results we obtained in the 2017 edition of the Emotion Recognition in the Wild Challenge \cite{dhall2017EmotiW}.   

Emotion recognition has received a lot of attention in the scientific literature. One large part of this literature deals with the possible options for defining and representing emotions. If the use of discrete classes~\cite{plutchik2013theories} such as joy, fear, angriness, \etc is the most straightforward way to do it, it can be more interesting to represent emotions by their degrees of arousal and valence, as proposed in~\cite{barrett1999structure}. In the restricted case of facial expressions, action units can also be used, focusing on the activation of different parts of the face~\cite{ekman1977facial}. Links can be made between these two representations: the discrete classes can be mapped into the arousal valence space~\cite{kossaifi2017afew} and can be deduced from the action units~\cite{khorrami2015deep}.

Another important part of the literature focuses on the ways to represent audio/video contents by features that can be subsequently used by classifiers. Early papers make use of (i) hand-crafted features such as Local Binary Patterns (LBP), Gabor features, discrete cosine transform for representing images, and Linear Predictive Coding coefficients (LPC),  relative spectral transform - linear perceptual prediction (RASTA-PLP), modulation spectrum (ModSpec) or Enhanced AutoCorrelation (EAC) for audio, (ii) standard classifiers such as SVN or KNN for classification (see \cite{kachele2016revisiting} for details). \cite{yao2015capturing}, the winner of EmotiW'15, demonstrated the relevance of Action Units for the recognition of emotions. \cite{baccouche2012spatio} was among the first to propose to learn the features instead of using hand-crafted descriptors, relying on Deep Convolutional Networks. More recently, \cite{fan2016video}, the winner of EmotiW'16, has introduced the C3D feature which is an efficient spatio-temporal representation of faces. 

The literature on emotion recognition from audio/video contents also  addresses the question of the fusion of the different modalities. A modality can be seen as one of the signals allowing to perceive the emotion. Among the recent methods for fusing several modalities, we can mention the use of  two-stream ConvNets \cite{Simonyan:2014ue},  ModDrop \cite{neverova2016moddrop} or  Multiple Kernel Fusion~\cite{chen2014emotion}. The most used modalities, in the context of emotion recognition, are face images and audio/speech, even if context seems also to be of tremendous importance~\cite{barrett2011context}. For instance, the general understanding of the scene, even based on simple features describing the whole image, may help to discriminate between two candidate classes. 

As most of the recent methods for emotion recognition are supervised and hence requires some training data, the availability of such resources is becoming more and more critical. Several challenges have collected useful data. The AVEC challenge~\cite{valstar2016avec} focuses on the use of several modalities to track arousal and valence in videos recorded in controlled conditions. The Emotionet challenge~\cite{benitez2017emotionet} proposes a dataset of one million images annotated with action units and partially with discrete compound emotion classes. Finally, the Emotion in the Wild challenge~\cite{dhall2016emotiw} deals with the classification of short video clips into seven discrete classes. The videos are extracted from movies and TV shows  recorded "in the wild". The ability to work with data recorded in realistic situations, including occlusions, poor illumination conditions, presence of several people or even scene breaks is indeed very important.

As aforementioned, this paper deals with our participation to the Emotion In the Wild 2017 (EmotiW) challenge.  We build on the state-of-the-art pipeline of \cite{fan2016video} in which i) audio features are extracted with the OpenSmile toolkit \cite{eyben2010opensmile}, ii) two video features are computed, one  by the C3D descriptor,  the other by the VGG16-Face model~\cite{parkhi2015deep} fine-tuned with FER2013 face emotion database and introduced into a LSTM network. Each one of these 3 features is processed by its own classifier, the output of the 3 classifiers being combined through late fusion. Starting from this pipeline, we propose to improve it in 3 different ways, which are the main contributions of our approach.

First, the recent literature suggests that late fusion might not be the optimal way to combine the different modalities (see \eg, \cite{neverova2016moddrop}). This paper investigates different directions, including an original hierarchical approach allowing  to combine scores (late fusion) and features (early fusion) at different levels. It can be seen as a way to combine information at its optimal level of description (from features to scores). This representation addresses an important issue of fusion, which is to ensure the preservation of unimodal information while being able to exploit cross-modal information.

Second, we investigate several ways to better use the temporal information in the visual descriptors. Among several contributions, we propose a novel descriptor combining C3D and LSTM.

Third, it can be observed that the amount of training data (773 labeled short video clips) is rather small compared to the number of the parameters of standard deep models, considering the complexity and diversity of emotions. In this context, supervised methods are prone to over-fitting. We show in the paper how the effect of over-fitting can be reduced by carefully choosing the number of parameters of the model and favoring transfer learning whenever it is possible.    

The rest of the paper is organized as follows: a presentation of the proposed model is done in Section \ref{overview}, detailing the different modalities and the fusion methods. Then Section \ref{results} presents the experimental validation as well as the results obtained on the validation and test sets during the challenge.

\begin{figure}[!tb]
\includegraphics[width=\linewidth]{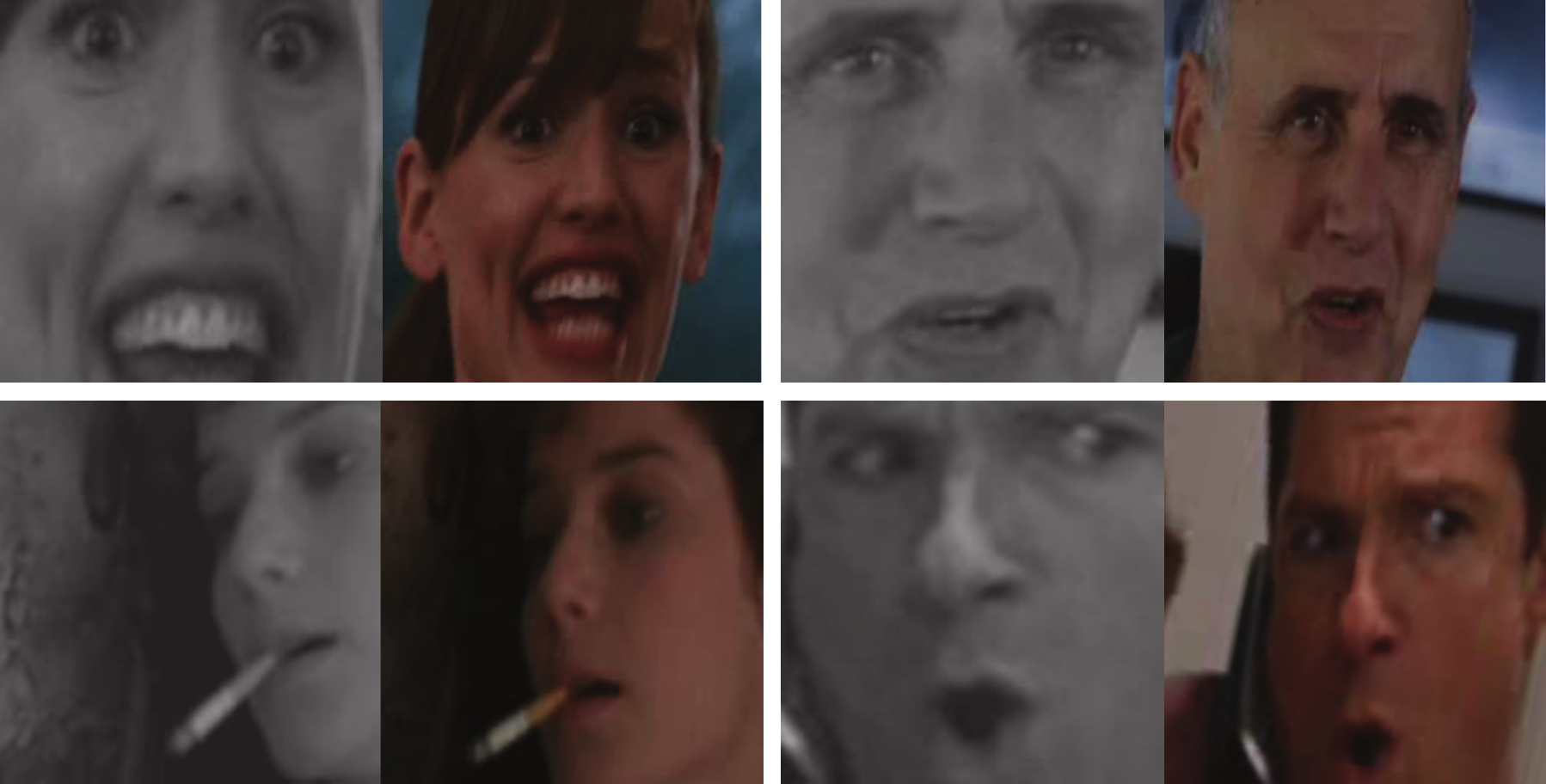}
\caption{Comparison of the bounding boxes given by our detection/normalization procedure (color images) with those provided with the AFEW dataset (gray-scale images), on 4 random faces of AFEW. Our bounding boxes are slightly larger.}
\label{Crop}
\end{figure}

\section{Presentation of the Proposed Approach \label{overview}}

\begin{figure*}[!tb]
\includegraphics[width=\linewidth]{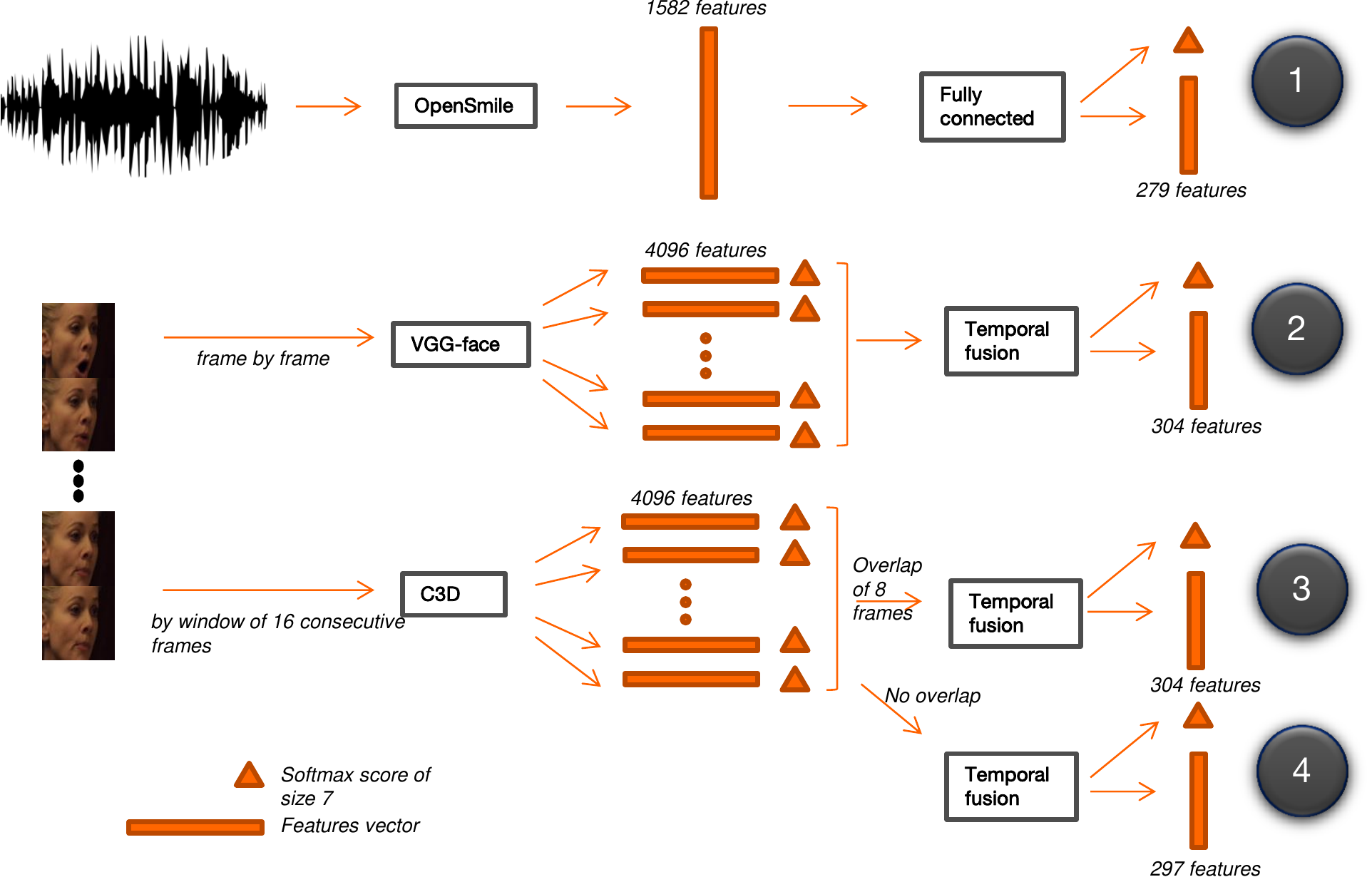}
\caption{Our model includes audio, VGG-LSTM and C3D-LSTM (with or without overlapping) modalities. Each modality gives its own feature vectors and scores, which can be fused using different methods. It can be noted that the dimensions of the features vectors are chosen to have balanced contributions between modalities, making a trade off between the number of parameters and the performance.}
\label{whole}
\end{figure*}

Figure \ref{whole} presents an overview of our approach, which is inspired by the one of \cite{fan2016video}.  Modalities we consider are extracted from audio and video, associated with faces analyzed with two different models (2D CNN, C3D). The two main contributions are the temporal fusion and the novel C3D/LSTM descriptor.

On overall, our method works as follows: on the one hand, the OpenSmile library \cite{eyben2010opensmile} is used to produce 1582 dimensional features used by a two-layer perceptron to predict classes as well as compact descriptors (279-d vectors) from audio. On the other hand, video classification is based on face analysis. After detecting the faces and normalizing their appearance, one set of features is computed with the VGG-16 model \cite{parkhi2015deep} while another set of features is obtained by the C3D model \cite{fan2016video}. In both cases, 4096-d features are produced. Temporal fusion of these so-obtained descriptions is done across the frames of the sequences, producing per modality scores and compact descriptors (304 dimensional vectors). The 3 modalities (audio, VGG-faces and C3D-faces) are then combined using both the score predictions and the compact representations.

After explaining how the faces are detected and normalized, the rest of the section gives more details on how each modality is processed, and how the fusion is performed.

\subsection{Face Detection and Alignment \label{sec:facedetection}}
The EmotiW challenge provides face detections for each frame of each video. However, we preferred not to use these annotations but to detect the faces ourselves.  The motivation is twofold. First, we want to be able to process any given video and not only those of EmotiW (\eg, for adding external training data).  Second, it is necessary to master the face alignment process for processing the images of face datasets when pre-training the VGG models (see Section~\ref{VGG} where FER dataset is used to pre-train the VGG model). 

For face detection, we use an internal detector provided by Orange Labs. We found out from our observations that this face detector detects more faces (20597 versus 19845 on the validation set) while having a lower false positive rate (179 false positives versus 908 on the validation set) than the one use to provide EmotiW annotations.

This detector is applied frame per frame. If one or several faces are detected, a tracking based on their relative positions allows to define several face tracks. In a second time, an alignment based on the landmarks ~\cite{king2009dlib} is done. We choose the longest face sequence in each video and finally apply a temporal smoothing of the positions of the faces to filter out jittering.

Figure~\ref{Crop} compares one of our normalized faces (after detection / alignment) with the one provided with the challenge data.

\subsection{Audio features and classifier}
The audio channel of each video is fed into the OpenSmile toolkit~\cite{eyben2010opensmile}\footnote{We use the {\tt emobase2010} configuration file.}, as most of the EmotiW competitors~\cite{dhall2016emotiw}, to get a description vector of length 1582.

A commonly used approach for audio is then to learn a classifier on top of the OpenSmile features. Support Vector Machine~\cite{fan2016video}~\cite{yao2015capturing} seems to be the dominant choice for the classification, even if there are some other approaches like Random Forests~\cite{fan2017spatiotemporal}.

To be able to control more finely the dimensionality of the  description vector (the one used later on during the fusion process), we learn a two-layer Perceptron with reLu activation on the OpenSmile description, using batch normalisation and dropout. During inference, we extract for each video a description vector of size 279 -- the hidden layer of the perceptron -- along with the softmax score. 

\subsection{Representing Faces with 2D Convolutional Neural Network \label{VGG}}

A current popular method (see \eg, \cite{fan2016video,kaya2017video}) for representing faces in the context of emotion recognition, especially in the EmotiW challenge, is to fine-tune a pre-trained 2D CNN such as the VGG-face model~\cite{parkhi2015deep} on the images of an emotion images dataset (\eg, the FER 2013 dataset). Using a pre-trained model is a way to balance the relatively small size of the EmotiW dataset (AFEW dataset). The images of the FER 2013 dataset~\cite{goodfellow2013challenges} are first processed by detecting and aligning the faces, following the procedure explained in Section~\ref{sec:facedetection}. We then fine-tune the VGG-face model on FER 2013 dataset, using both the training and the public test set;  during training we use data augmentation  by jittering the scale, flipping and rotating the faces. The aim is to make the network more robust to small misalignment of the faces. We also apply a strong dropout on the last layer of the VGG (keeping only 5\% of the nodes) to prevent over-fitting. We achieve a performance of 71.2\% on the FER private test set, which is slightly higher than the previously published results~\cite{fan2016video,goodfellow2013challenges}.

To assess the quality of the description given by this fine-tuned VGG model to emotion recognition, we first benchmarked it on the validation set of SFEW~\cite{dhall2015video}, which is a dataset containing frames of the videos of AFEW (those used by the EmotiW challenge). We achieved a score of 45.2\% without retraining the model on SFEW, while the state-of-the-art result~\cite{kim2016hierarchical} is 52.5\%, using a committee of deep models, and the challenge baseline is of 39.7\%~\cite{dhall2015video}.

The face sequences detected in AFEW videos are resized to 224x224 and fed into the fine-tuned VGG-face model. We then extract the 4096 length fc6 layer, following the same pipeline as Fan et al.~\cite{fan2016video,fan2017spatiotemporal}. We also compute the softmax score for each frame.

\subsection{Representing Faces with 3D Convolutional Neural Network}

3D convolutional neural networks have been shown to give good performance in the context of facial expressions recognition in  video~\cite{tran2015learning,baccouche2011sequential}. Fan \etal~\cite{fan2016video} fine-tuned a pre-trained 'sport1m' model on randomly chosen windows of 16 consecutive faces. During inference, the model is applied to the central frame of the video.

%We first applied the same pipeline and did not reach the same performances without over-fitting on the training set. A model pre-trained on sport1m and fine-tuned on UCF101~\cite{soomro2012ucf101} was also tested, bringing no improvement.
%
%A second try was done by picking middle window for both the training and the validation. We also applied weight decay and dropout to prevent over-fitting.

One limitation of this approach is that, at test time, there is no guaranty that the best window for capturing the emotion is in the middle of the video. The same problem occurs during training: a large part of the windows (randomly) selected for training does not contain any emotion or does not contain the correct emotion. Indeed, videos are annotated as a whole but some frames can have different labels or, sometimes, no expression at all.

We address this limitation in the following way: we first fine-tune a C3D-sport1m model using all the windows of each video, optimizing the classification performance until the beginning of convergence. Then, to be able to learn more from the most meaningful windows than from the others, we weigh each window based on its scores. More precisely,
for the $i^{th}$ video and the $j^{th}$ window of this video, at epoch $t$, the weight $w_{i,j}$ is computed as:
\begin{equation}
  w_{i,j}=e^{\frac{-s_{i,j}}{T(t)}} \label{eq:w}
\end{equation}
where $T(t)$ is a temperature parameter decreasing with epoch $t$
and $s_{i,j}$, the score of the window $j$ of the video $i$.
We then normalize the weight to ensure that for each video $i$, $\sum_{j}w_{i,j}=1$. A random grid search on the hyper-parameters, including the temperature descent, is made on the validation set.

\begin{figure}[tb]
\includegraphics[width=\linewidth]{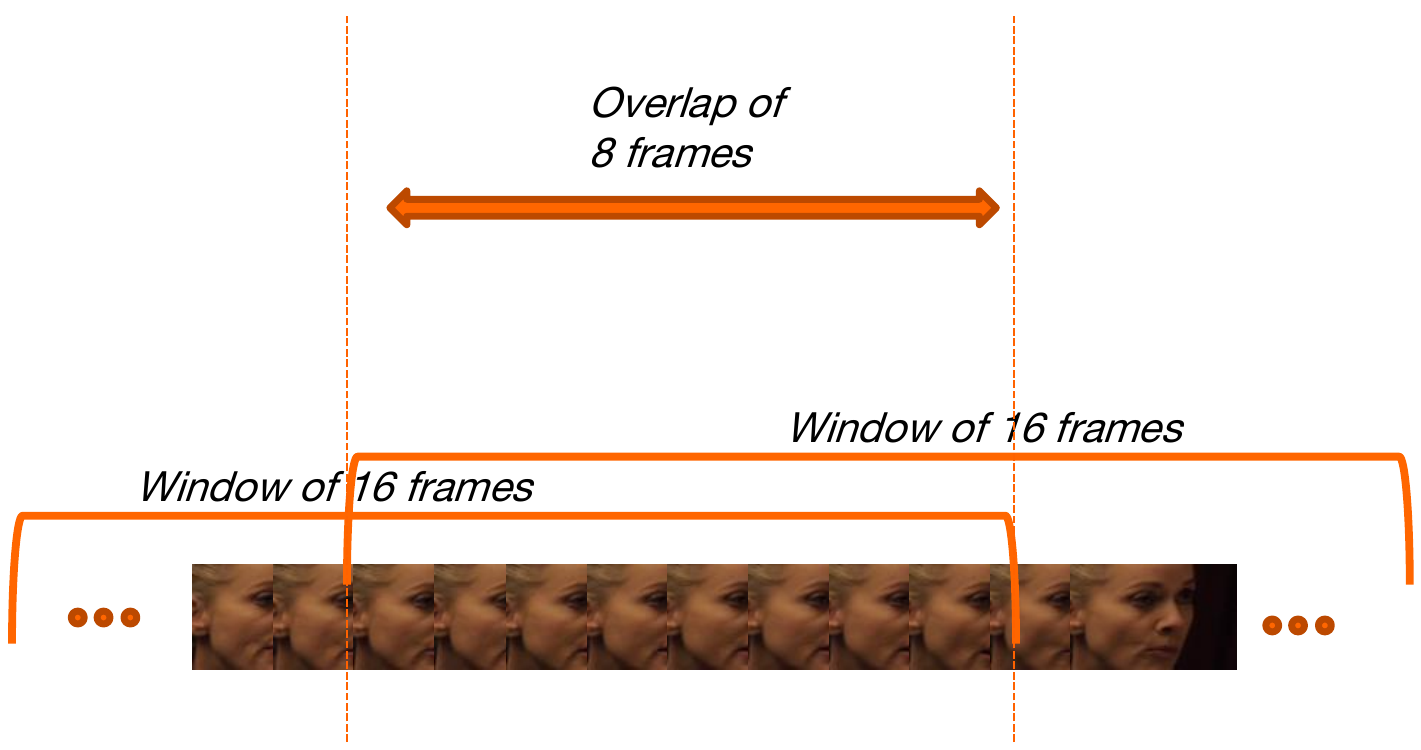}
\caption{In the case of 8 frames overlap, we can see that the 16-frame windows are sharing half of their faces.}
\label{overlap}
\end{figure}

During inference, the face sequences detected in AFEW videos are resized to 112x112. We then split a video into several windows of 16 consecutive frames\footnote{The sequences with fewer than 16 frames were padded by themselves to reach a sufficient length.}, with or without overlapping between the windows (as shown in Figure~\ref{overlap}), and fed it into the weighted C3D. We then extract the second 4096-d  fully connected layer and the last softmax layer of the C3D for each window of the videos.

The method we propose bears similarities with Multiple Instance Learning (MIL)~\cite{ali2010human}. Within the framework of MIL, each video can be considered as a bag of windows, with one single label. A straightforward way to apply MIL would be to train the C3D on each video as a bag of windows, and add a final layer for choosing the prediction as the one with the maximum score among all the scores of the batch. The loss would be then computed from this prediction. The weights defined in Eq.(\ref{eq:w}) play this role by selecting iteratively the best scoring windows. 

\subsection{Per Modality Temporal Fusion}
Both VGG and weighted C3D representations are applied to each frame of the videos,  turning the videos into temporal sequences of visual descriptors. We will name the elements of those sequences as "descriptors", whether it is the description of a frame or of a window.

To classify these sequences of descriptors, we investigated several methods. The most straightforward one is to score each descriptor and to take the maximum of the softmax scores as the final prediction. Similarly, the maximum of the means of the softmax across time can also be considered.

To better take into account temporal dependencies between consecutive descriptors, another option is to use Long Short-Term Memory recurrent neural networks (LSTM)~\cite{baccouche2011sequential,gers1999learning}. Unlike \cite{fan2016video,fan2017spatiotemporal}, we chose to use a variable length LSTM allowing us to take all the descriptors as inputs. 

To prevent over-fitting, we also applied dropout to the LSTM cells and weight decay to the final fully connected layer. A random grid search on hyper-parameters is then applied for each one of these models.

Both VGG-LSTM and C3D-LSTM are used to give one description vector (output of the final fully connected layer of the LSTM) and one softmax score for each video.

The final VGG-LSTM architecture has 2230 hidden units for each LSTM-cell and a final fully connected layer with 297 hidden units. The maximal length of the input sequence is of 272 frames.

The final C3D-LSTM architecture has 1324 hidden units for each LSTM-cell and a final fully connected layer of 304 hidden units. The maximal length of the input sequence is of 34 windows (overlap of 8 frames) or 17 windows (no overlap).

\subsection{Multimodal fusion}
Last but not the least, the different modalities have to be efficiently combined to maximize the overall performance. 

As explained in Section~\ref{sec:intro}, two main fusion strategies can be used \ie score fusion (late fusion), which consists in predicting the labels based on the predictions given by each modality, or  features fusion (early fusion), which consists in taking as input the latent features vectors given by each modality and learning a classifier on the top of them.

During the last editions of the challenge, most of the papers focused on score fusion, using SVM~\cite{bargal2016emotion}, Multiple Kernel Fusion~\cite{chen2014emotion} or weighted means~\cite{fan2016video}. Differently, several authors tried to train audio and image modalities together (see \eg,~\cite{chao2016audio}), combining early features and using soft attention mechanism, but they didn't achieve state-of-the-art performance. We propose an approach combining both.

The rest of the section describes the four different fusion methods we experimented with, including a novel method (denominated as {\em score trees}).

\paragraph{Baseline score fusion}
We experimented with several standard score fusion, like majority voting, means of the scores, maximum of the scores and linear SVM.

\begin{figure}[tb]
\includegraphics[width=\linewidth]{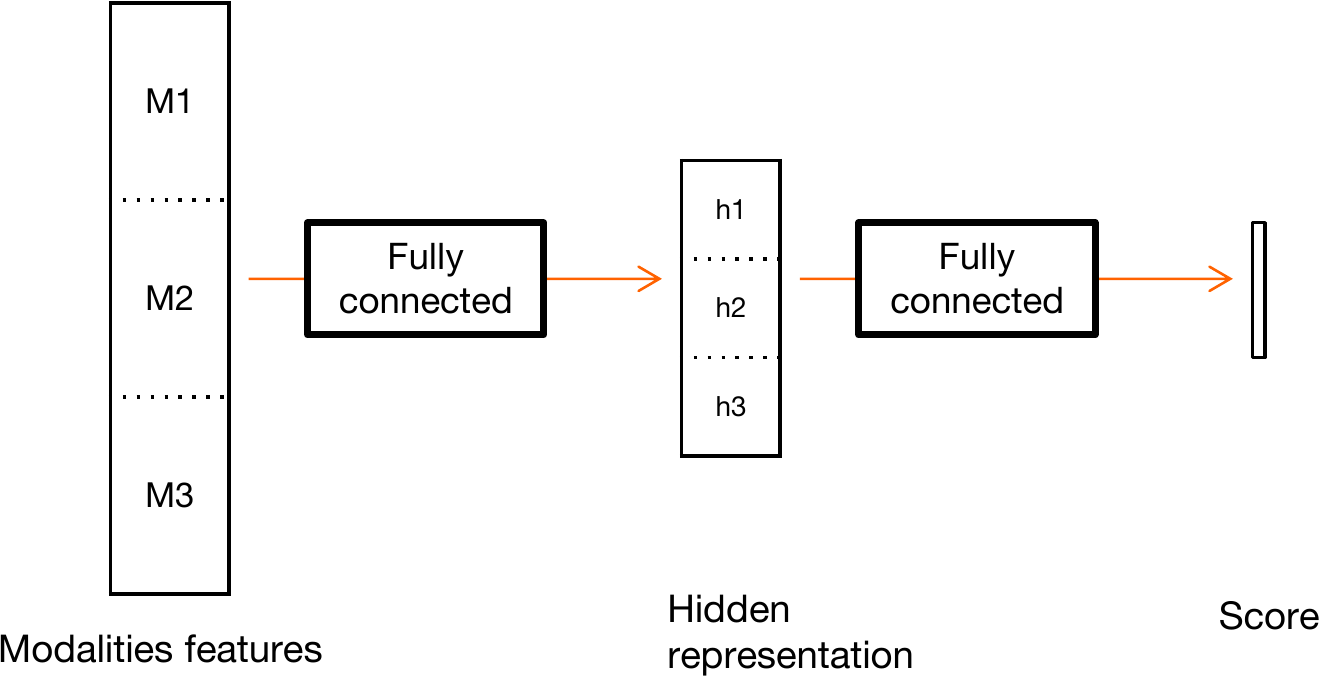}
\caption{Fully connected model with ModDrop. The three modalities are concatenated and a fully connected with modDrop is applied. The obtained hidden representation is then fed to a second regular fully connected layer, which will output the scores.}
\label{modDrop}
\end{figure}

\paragraph{Fully connected neural network with modality drop}
We also experimented with the ModDrop method of Neverova \etal~\cite{neverova2016moddrop}. It consists in applying dropout by modality, learning the cross-modality correlations while keeping unimodal information.  \cite{neverova2016moddrop} reports state-of-the-art results on gesture recognition. We apply this method to our audio, C3D-LSTM and VGG-LSTM features, as shown in Figure~\ref{modDrop}. According to \cite{neverova2016moddrop}, this is much better than simply feeding the concatenation of the modalities features into a fully connected neural network and letting the network learn a joint representation. Indeed, the fully connected model would be unable to preserve unimodal information during cross-modality learning. 

An important step to make convergence possible with ModDrop is to first learn the fusion without cross-modality. For this reason, \cite{neverova2016moddrop} conditioned the weight matrix of the first layer so that the diagonal blocks are equal to zeros and released this constraint after a well-chosen number of iterations.

To warranty the preservation of the unimodal information, we explore an alternative method which turned out to be better: we apply an adapted weight decay, only on the non-diagonal blocks, and decreased its contribution to the loss through time. 

To be more formal, let $n$ be the number of modalities, $W1$ be the weights matrix of the first layer. It can be divided into $n^{2}$ weights block matrices $W1_{k,l}$, modeling unimodal and intermodal contributions, with $k$ and $l$, ranging over the number of modalities $n$.
The first fully connected equation can be written as :
\begin{displaymath}
	\begin{pmatrix}
    h_{1} \\ h_{2} \\ h_{3}\\
    \end{pmatrix}
    =
	\begin{pmatrix}
	W1_{1,1} & W1_{1,2} & W1_{1,3} \\
	W1_{2,1} & W1_{2,2} & W1_{2,3} \\
	W1_{3,1} & W1_{3,2} & W1_{3,3} \\
	\end{pmatrix}
    \begin{pmatrix}
	M_{1} \\ M_{2} \\ M_{3} \\
	\end{pmatrix}
\end{displaymath} 
Then, the term to add to the loss is simply (with $\gamma_{md}$ decreasing through the time):
\begin{displaymath}
	\gamma_{md} \sum_{k\neq l}||W1_{k,l}||_{2}
\end{displaymath}

Setting $\gamma_{SD}$ to very high values during the first iterations leads to zeroing non-diagonal block matrices. Lowering it later reintroduces progressively these coefficients. From our observation, this approach provided better convergence on the considered problem.

\paragraph{Score trees}

\begin{figure}[tb]
\includegraphics[width=\linewidth]{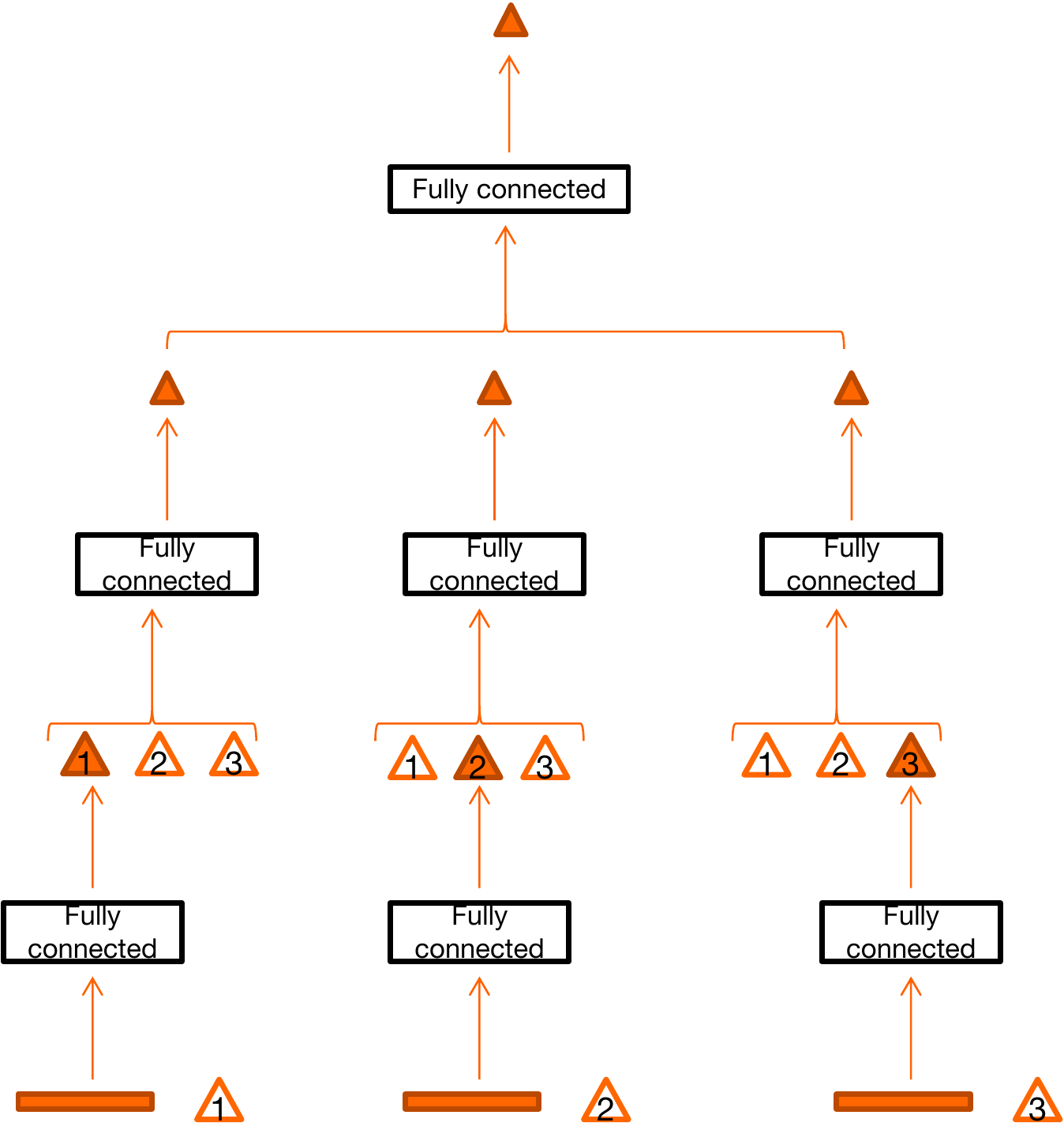}
\caption{The {\em Score Tree} architecture}
\label{scoreTree}
\end{figure}

Our motivation is to combine the high-level information coming from the scores with the lower-level information coming from the features,  all together. We did it by building what we call {\em Score Trees} (see Figure~\ref{scoreTree} for an illustration). 

A fully connected classification neural network is applied separately to the features of the different modalities, outputting a vector of size 7. This vector is then concatenated with the scores of the two other modalities, to create a vector of size 21. A fully connected classification neural network is then fed with it and outputs a prediction vector of size 7. The aim is to make predictions with respect to the predictions coming from other modalities. Finally, these three new prediction vectors are concatenated and fed into a last fully connected classifier, which gives the overall scores. This method can be generalized to any number of modalities.

\paragraph{Weighted Mean}
The weighted mean is the approach of the winners of the 2016 edition~\cite{fan2016video}. It consists in weighting the score of each modality and sum them up.

The weights are chosen by cross validation on the validation set, selecting the ones giving the best performance.

We applied it on the VGG-LSTM, C3D-LSTM and audio models.

%\subsection{Implementation details}
%The internal face detector is implemented with Torch7.
%The audio features were extracted with OpenSmile, using the emobase2010 configuration file.
%The rest of the pipeline was implemented with Tensorflow 1.1.

\section{Experimental Validation and Results \label{results}}

After introducing the AFEW dataset (the dataset of the challenge), this section presents the experimental validation of our method. We first present experiments done on the modalities taken separately, and then present experiments on their fusion. We finally introduce the experiments done for the EmotiW'17 challenge and the performance we obtained. 

\subsection{The Acted Facial Emotion in the Wild dataset}
Acted Facial Emotion in the Wild (AFEW) is the dataset used by the EmotiW challenge. The 2017 version of AFEW is composed of 773 training videos, 383 validation videos and 653 test videos. Each video is labeled with one emotion among: 'angry', 'disgust', 'fear', 'happy', 'sad', 'neutral' and 'surprise'. In addition to the video, cropped and aligned faces extracted by ~\cite{zhu2012face,xiong2013supervised} are also provided. 

\begin{table}[tb]
\centering
\begin{tabular}{p{0.25\linewidth}|p{0.2\linewidth}|p{0.2\linewidth}|p{0.2\linewidth}|}
\cline{2-4}
\textbf{}                               & \textbf{Training} & \textbf{Validation} & \textbf{Test} \\ \hline
\multicolumn{1}{|c|}{\textbf{Angry}}    & 133 (17.2 \%)     & 64 (16.7 \%)        & 99 (15.2 \%)  \\ \hline
\multicolumn{1}{|c|}{\textbf{Disgust}}  & 74 (9.6 \%)       & 40 (10.4 \%)        & 40 (6.1 \%)   \\ \hline
\multicolumn{1}{|c|}{\textbf{Fear}}     & 81 (10.4 \%)      & 46 (12 \%)          & 70 (10.7 \%)  \\ \hline
\multicolumn{1}{|c|}{\textbf{Happy}}    & 150 (19.4 \%)     & 63 (16.4 \%)        & 144 (22 \%)   \\ \hline
\multicolumn{1}{|c|}{\textbf{Sad}}      & 117 (15.1 \%)     & 61 (15.9 \%)        & 80 (12.3 \%)  \\ \hline
\multicolumn{1}{|c|}{\textbf{Neutral}}  & 144  (18.6 \%)    & 63 (16.4 \%)        & 191 (29.2 \%) \\ \hline
\multicolumn{1}{|c|}{\textbf{Surprise}} & 74 (9.6 \%)       & 46 (12 \%)          & 29 (4.4 \%)   \\ \hline
\multicolumn{1}{|c|}{\textbf{Total}}    & 773               & 383                 & 653           \\ \hline
\end{tabular}
\caption{AFEW 7.0: number of video sequences per class.}
\label{distrib}
\end{table}

Another important specificity of this dataset is  the class distribution of the Train/Val/Test subsets (as shown in Table \ref{distrib}). This difference can make the performance on the Test set different from the one of the Validation set, as some classes are more challenging than others.

\subsubsection{External training data}
To enlarge the training set, we collected external data by selecting 380 video clips from our personal DVDs movie collection, after checking that there is no overlap between the selected movies and the ones of AFEW~\cite{dhall2012collecting}. These movies were processed by the following pipeline: faces are first detected using our face detector (see Section~\ref{sec:facedetection} for details), and the bounding boxes and timestamps are kept. We then extract candidates temporal windows of ten seconds around all of these time stamps and asked human annotators to select and annotate the most relevant ones.
To ensure the quality of our annotations, we evaluated ourselves (as human beings) on the validation set. We reached a performance from 60~\% to 80~\% depending on the annotator, which is compatible with the figure of 60\% observed by~\cite{kachele2016revisiting}. 

\subsection{Experiments on Single Modalities}
Each modality has been evaluated separately on the validation set of AFEW. The VGG-LSTM and C3D-LSTM modalities performs better than~\cite{dhall2016emotiw}.

Regarding the VGG-LSTM and the C3D-LSTM, both unidirectional and bidirectional LSTM architectures~\cite{graves2005bidirectional} were evaluated, with one or two layers. 

In several recent  papers~\cite{baccouche2012spatio,du2015hierarchical}, bidirectional LSTM is claimed to be more efficient than the unidirectional one and could be seen as a way to augment the data. Nevertheless, in our case, we have observed that the bidirectional LSTM was prone to over-fitting on the training set and therefore does not perform well on the validation set. The same observation has been made when increasing the number of layers. 
The best performing architecture is in both cases a one-layer unidirectional LSTM.

\subsubsection{VGG-LSTM}
The VGG model (without LSTM) has first been evaluated by taking the maximum of the scores over the sequences, giving the accuracy of 41.4\% on the validation set.

Then the different LSTM architectures were tested and the best performance for each one is given in Table \ref{VGGLSTM}. We note a 3\% improvement compared to Fan \etal~\cite{fan2016video}. It can be explained by the fact that our model uses the whole sequences, feeding the model with more information. Data augmentation also helps.

\begin{table}[tb]
\centering
\begin{tabular}{|p{0.65\linewidth}|p{0.25\linewidth}|}
\hline
\textbf{Method}                         & \textbf{Validation accuracy} \\ \hline
Maximum of the scores                   & 41.4 \%                      \\ \hline
\textbf{Unidirectional LSTM  one layer} & \textbf{48.6 \%}             \\ \hline
Bidirectional LSTM one layer            & 46.7 \%                      \\ \hline
Unidirectional LSTM two layers          & 46.2 \%                      \\ \hline
Bidirectional LSTM two layers           & 45.2 \%                      \\ \hline
 {Fan \etal~\cite{fan2016video}} 			&  {45.42 \%}             \\ \hline
\end{tabular}
\caption{VGG-LSTM performance on the validation set of AFEW.}
\label{VGGLSTM}
\end{table}

\subsubsection{C3D-LSTM}
We then experimented our method with the C3D modality alone. The performance is given in Table~\ref{C3D}.

We observe that our implementation of C3D trained on random windows and evaluated on central windows is not as good as Fan \etal~\cite{fan2016video}. The C3D trained on central windows performed better but is not state-of-the-art either.

Our proposed C3D (with LSTM) has been tested with and without overlapping between windows. To evaluate the weighted C3D, the prediction of the window with the maximal softmax score among the video is first taken. It performs better without overlapping, and we observe a lower difference between training and validation accuracy. It could be explained by the fact that the number of windows is lower if there is no overlap, the choice between the windows is therefore easier. As a second observation, the use of LSTM with the weighted C3D leads to the highest scores.

At the end, it can be observed that our C3D descriptor performs significantly better than the one of~\cite{fan2016video}.

\begin{table}[tb]
\centering
\begin{tabular}{|p{0.65\linewidth}|p{0.25\linewidth}|}
\hline
\textbf{Method}                      & \textbf{Validation accuracy} \\ \hline
C3D on central window               & 38.7 \%                      \\ \hline
C3D on random window                 & 34 \%                        \\ \hline
 {Weighted C3D (no overlap)}   &  {42.1 \%}             \\ \hline
Weighted C3D (8 frames overlap)      & 40.5 \%                      \\ \hline
Weighted C3D (15 frames overlap) 	& 40.1 \%             			\\ \hline
\textbf{LSTM C3D (no overlap)}       & \textbf{43.2 \%}             \\ \hline
 {LSTM C3D (8 frames overlap)} &  {41.7 \%}             \\ \hline
Fan \etal~\cite{fan2016video} 			&  {39.7 \%}           \\ \hline

\end{tabular}
\caption{C3D and C3D-LSTM performance on the validation set of AFEW.}
\label{C3D}
\end{table}

\subsubsection{Audio}

The audio modality gave a performance of 36.5\%, lower than the state-of-the-art method (39.8\%)~\cite{fan2017spatiotemporal}. The use of a perceptron classifier (worse than the SVM) nevertheless allowed us to use high-level audio features during fusion.

\subsection{Experiments on Fusion}
Table~\ref{fusion} summarizes the different experiments we made on fusion.

The simple baseline fusion strategy (majority vote or means of the scores) does not perform as well as the VGG-LSTM modality alone.

The proposed methods (ModDrop and Score Tree) achieved promising results on the validation set, but are not as good as the simple weighted mean on the test set. This can be explained by the largest number of parameters used for the modDrop and the score tree, and by the fact that some parameters cross validated on the validation set.

The best performance obtained on the validation set has the accuracy of 52.2~\% , which is significantly higher than the performance of the baseline algorithm provided by the organizers -- based on computing LBPTOP descriptor and using a SVR -- giving the accuracy of 38.81~\% on the validation set \cite{dhall2017EmotiW}.

\begin{table}[tb]
\centering
\begin{tabular}{|p{0.5\linewidth}|p{0.2\linewidth}|p{0.15\linewidth}|}
\hline
\textbf{Fusion method} & \textbf{Validation accuracy} & \textbf{Test accuracy} \\ \hline
Majority vote          & 49.3 \%                        & \_                     \\ \hline
Mean                   & 47.8 \%                      & \_                     \\ \hline
ModDrop (sub.3)                & \textbf{52.2 \%}             & 56.66 \%               \\ \hline
Score Tree (sub.4)             & 50.8 \%                      & 54.36 \%               \\ \hline
Weighted mean (sub.6)         & 50.6 \%                      & \textbf{57.58 \%}      \\ \hline
\end{tabular}
\caption{Performance of the different fusion methods on the validation and test sets.}
\label{fusion}
\end{table}

\subsection{Our participation to the EmotiW'17 challenge}
We submitted the method presented in this paper to the EmotiW'17 challenge. 
We submitted 7 runs which performance is given in Table \ref{submissions}). 

The difference between the runs are as follows:
\begin{itemize}
\item Submission 2: ModDrop fusion of audio, VGG-LSTM and a LSTM-C3D with an 8-frame overlap.

\item Submission 3: addition of another LSTM-C3D, with no overlap, improving the performance on the test set as well as on the validation set.

\item Submission 4: fusion based on Score Trees, did not achieve a better accuracy on the test set, while observing a slight improvement on the validation set.

\item Submission 5: addition of one VGG-LSTM and two other LSTM-C3D, one with 8 frames overlap and one without. These new models were selected among the best results in our random grid search on hyper-parameters according to their potential complementarity degree, evaluated by measuring the dissimilarity between their confusion matrices. The fusion method is ModDrop.

\item Submission 6: weighted mean fusion of all the preceding modalities, giving a gain of 1~\% on the test set, while losing one percent on the validation set, highlighting generalization issues.

\item Submission 7: our best submission, which is the same as the sixth for the method but with models trained on both training and validation sets. This improves the accuracy by 1.2 \%. This improvement was also observed in the former editions of the challenge. Surprisingly, adding our own data didn't bring significant improvement (gain of less than one percent on the validation set). This could be explained by the fact that our annotations are not correlated enough with the AFEW annotation. 
\end{itemize}

\begin{table}[tb]
\centering
\begin{tabular}{|p{0.4\linewidth}|p{0.5\linewidth}|}
\hline
\textbf{Submission} & \textbf{Test Accuracy} \\ \hline
2                   & 55.28 \%               \\ \hline
3                   & 56.66 \%               \\ \hline
4                   & 54.36 \%               \\ \hline
5                   & 56.51 \%               \\ \hline
6                   & 57.58 \%               \\ \hline
7                   & 58.81 \%               \\ \hline
\end{tabular}
\caption{Performance of our submissions on the test set.}
\label{submissions}
\end{table}

The proposed method has been ranked 4th in the competition. We observed that, this year, the improvement of the top accuracy compared to the previous editions is small (+1.1\%), while from 2015 to 2016 the improvement was of +5.2~\%. This might be explained by the fact that the methods are saturating, converging towards human performance (which is assumed to be around 60~\%). However, the performance of top human annotators (whose accuracy is higher than 70~\%) means there is still some room for improvement.

\begin{figure}
\includegraphics[width=\linewidth]{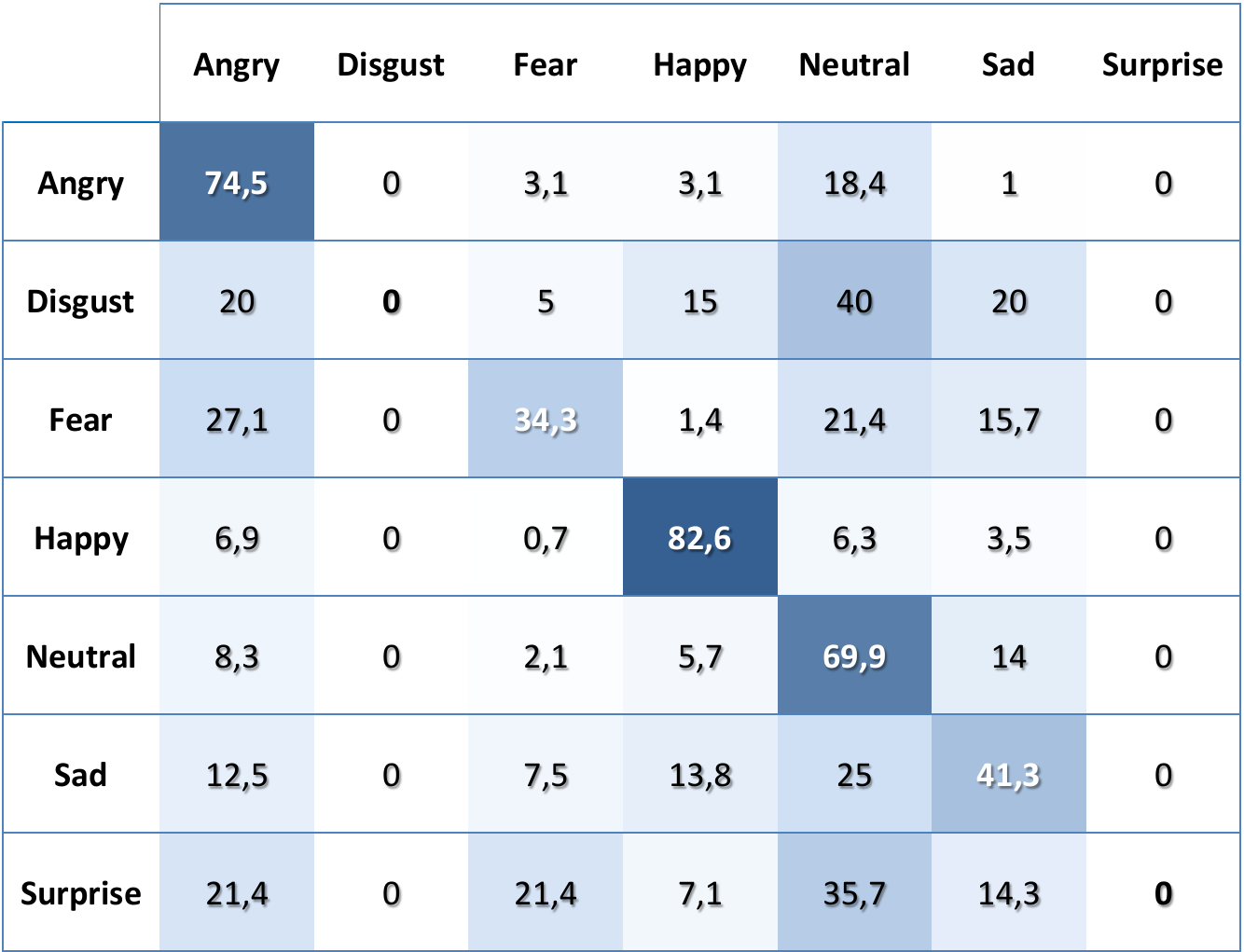}
\caption{Confusion matrix obtained with the seventh submission. We can see that 'disgust' and 'surprise' classes are never predicted by our model, while the three dominant classes ('happy', 'neutral' and 'angry') are well recognized. The 'neutral' class has the largest number of false positives. It underlines the difficulty, even for humans, to draw the margin between presence and absence of emotion. Rows denote true classes, columns predicted classes.}
\label{confMatrix}
\end{figure}

\section{Conclusions}
This paper proposes a multimodal approach for video emotion classification, combining VGG and C3D models as image descriptors and explores different temporal fusion architectures. Different multimodal fusion strategies have also been proposed and experimentally compared, both on the validation and on the test set of AFEW. At the EmotiW'17 challenge, the proposed method ranked 4th with the accuracy of \textbf{58.81~\%}, 1.5~\% under the competition winners. One important observation from this competition is the discrepancy between the performance obtained on the test set and the one on the validation set: good performance on the validation set is not a warranty to good performance on the test set. Reducing the number of parameters in our models could help to limit overfitting. Using pre-trained fusion models and, moreover, gathering a larger set of data would also be a good way to face this problem. Finally, another interesting path for future work would be to add  contextual information such as scene description, voice recognition or even movie type as an extra modality.
\bibliographystyle{ACM-Reference-Format}
\bibliography{bibliography} 

\end{document}